\documentclass{article}

    \PassOptionsToPackage{numbers, compress}{natbib}

    \usepackage[preprint]{neurips_2020}

\usepackage[utf8]{inputenc} 
\usepackage[T1]{fontenc}    
\usepackage{hyperref}       
\usepackage{url}            
\usepackage{booktabs}       
\usepackage{amsfonts}       
\usepackage{nicefrac}       
\usepackage{microtype}      

\usepackage{graphicx}
\usepackage{caption}
\usepackage{subcaption}
\usepackage{wrapfig}
\usepackage{caption}
\usepackage{amsmath}
\DeclareMathOperator{\sign}{sgn}
\usepackage{amsfonts}
\usepackage{amssymb}
\usepackage{tabularx}
\usepackage{amsthm}

\theoremstyle{definition}
\newtheorem{definition}{Definition}[section]

\title{Manifolds for Unsupervised Visual Anomaly Detection}

\author{%
  Louise Naud\\
  Augustus Intelligence\\
  New York City, NY 10014 \\
  \texttt{louise.naud@augustusai.com} \\
  \And
  Alexander Lavin \\
  Augustus Intelligence\\
  New York City, NY 10014 \\
  \texttt{alexander.lavin@augustusai.com} \\
}

\begin{document}

\maketitle

\begin{abstract}
Anomalies are by definition rare, thus labeled examples are very limited or nonexistent, and likely do not cover unforeseen scenarios. Unsupervised learning methods that don't necessarily encounter anomalies in training would be immensely useful. Generative vision models can be useful in this regard but do not sufficiently represent normal and abnormal data distributions. To this end, we propose constant curvature manifolds for embedding data distributions in unsupervised visual anomaly detection.
Through theoretical and empirical explorations of manifold shapes, we develop a novel hyperspherical Variational Auto-Encoder (VAE) via stereographic projections with a gyroplane layer - a complete equivalent to the Poincar{\'e} VAE. This approach with manifold projections is beneficial in terms of model generalization and can yield more interpretable representations. We present state-of-the-art results on visual anomaly benchmarks in precision manufacturing and inspection, demonstrating real-world utility in industrial AI scenarios. We further demonstrate the approach on the challenging problem of histopathology: our unsupervised approach effectively detects cancerous brain tissue from noisy whole-slide images, learning a smooth, latent organization of tissue types that provides an interpretable decisions tool for medical professionals.
\end{abstract}
\section{Introduction}



Annotating visual data can be burdensome and expensive in most real-world applications; for example, medical professionals manually inspecting and labeling massive whole-slide images (WSI) for thousands of nucleotides, lymphocytes, tumors, etc. This is exponentially so when trying to label a sufficient amount of anomalous data, as anomalies are by definition rare; even more, we have to assume there are unforeseen anomalous scenarios to arise in the future. Unsupervised methods are thus advantageous, and have seen promising advances with deep generative vision models. Recent and noteworthy work has been developing methods with Variational Auto-Encoders (VAE) \cite{Kingma2014AutoEncodingVB, Rezende2014StochasticBA} and Generative Adverserial Networks (GAN) \cite{Goodfellow2014GenerativeAN} towards these tasks \cite{An2015VariationalAB, Zong2018DeepAG, Schlegl2017UnsupervisedAD, Pidhorskyi2018GenerativePN, Deecke2018ImageAD}. 

Deep generative models learn a mapping from a low-dimensional latent space to a high-dimensional data space, centered around \textit{the manifold hypothesis}: high-dimensional observations are concentrated around a manifold of much lower dimensionality. It follows that by learning the proper manifold we can model the observed data with high-fidelity.
It is our aim to investigate properties of nonlinear manifolds and regularity conditions that behoove visual data representation for anomaly detection. We hypothesize Riemannian manifold curvatures other than the typical flat, Euclidean space can provide a more natural embedding on which to infer anomalous data in images.

Non-Euclidean latent spaces have recently been proposed in deep generative models, namely hyperbolic and hyperspherical metric spaces. With the former, the latent Poincar{\'e} space is shown to learn hierarchical representations from textual and graph-structured data \cite{Nickel2017PoincarEF, tifrea2018poincare}, and from images with the Poincar{\'e} VAE of \citet{Mathieu2019ContinuousHR}. 
Spherical embedding spaces have been shown useful for class separation and smooth interpolation in the manifold towards computer vision tasks \cite{Mettes2019HypersphericalPN, Davidson2018HypersphericalVA, Haney2020HypersphereVision}.
We hypothesize these manifold geometries naturally represent distinct normal and abnormal visual data distributions, and can be learnt from data without labels via latent manifold mappings in deep generative models. We take care to investigate the properties of these manifolds most relevant to learning and inferring on unlabeled visual data, and carry out thorough  experiments to understand the effects of various Riemannian manifold regimes. We indeed confirm our hypotheses and develop novel VAE methods for utilizing the various manifold curvatures.


\textbf{Our main contributions}\footnote{Code will be open-sourced with camera-ready publication.}:
\begin{enumerate}
    \item Theoretical utilities of Riemannian manifolds for the generative model latent space, towards naturally and efficiently embedding both normal data and sparse anomalous data.
    \item Proposal of \textit{Stereographic Projection Variational Auto-Encoders}, towards unsupervised visual anomaly detection. We derive a novel \textit{gyroplane layer} for a neural network to be capable of stereographic projections across hyperspherical and hyperbolic manifold shapes.
    \item Empirical analyses of our approach vs comparable methods on challenging benchmark datasets for unsupervised visual anomaly detection, achieving state-of-the-art results.
    \item Neuropathology experiments that show our VAE method can reliably organize the various subtypes of brain tissue without labels, and identify anomalous tissues samples as cancerous. We further motivate the hyperbolic latent space by demonstrating \textit{Poincar{\'e} mapping}, to  visualize the latent organization and reliably interpolate between regions of normal and abnormal brain tissue.
\end{enumerate}

\section{Representation Learning in Generative Vision}


\begin{figure}[!ht]
  \centering
  {\includegraphics[width=0.75\linewidth]{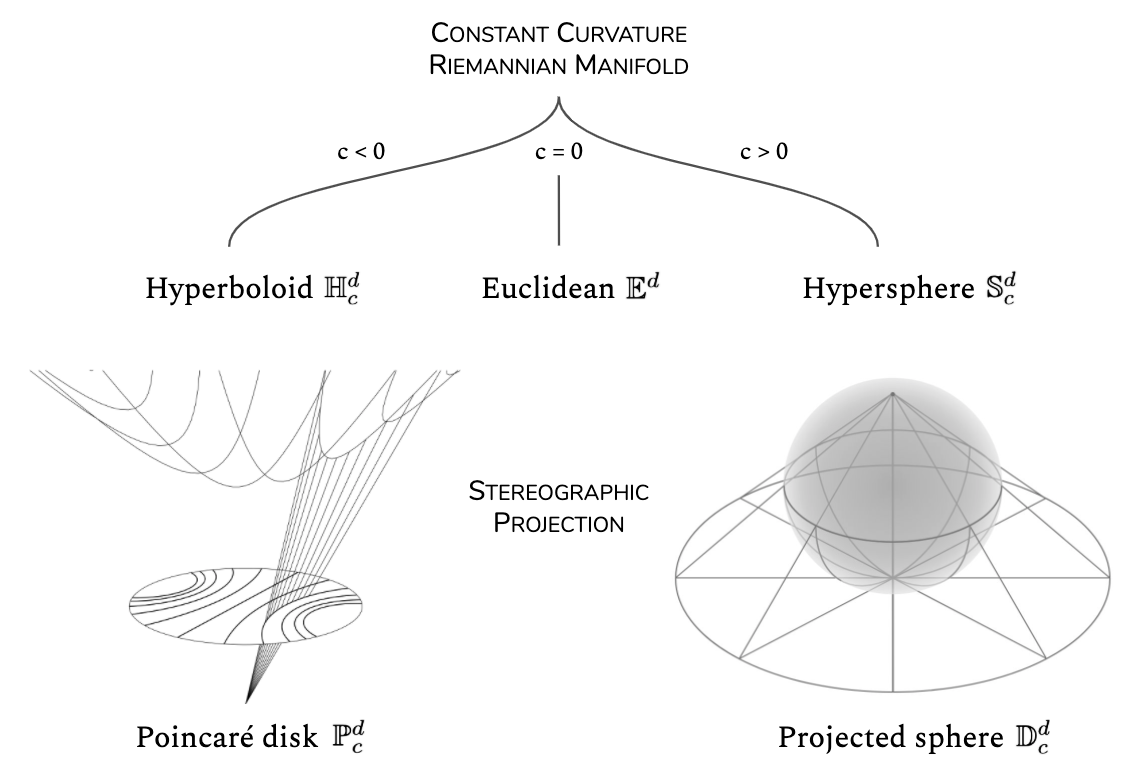}}
  \caption{The three regimes of constant curvature Riemannian manifolds, for which we can utilize the stereographic projections of the hyperboloid $\mathbb{H}_{c}^{d}$ (left) and hypersphere $\mathbb{S}_{c}^{d}$ (right) to respectively yield the Poincar{\'e} ball $\mathbb{P}_{c}^{d}$ and projected sphere $\mathbb{D}_{c}^{d}$ manifolds. Example geodesic arcs of this projection are shown. The mapping is smooth, bijective, and conformal (preserving the angles at which curves meet). This projection is necessary to yield manifolds with consistent modeling properties across the spectrum of curvatures $c$ (see text for details).}
  \label{fig:stereo_projections}
\end{figure}






\subsection{Properties of Manifold Curvatures}

Consider a true data-generating process that draws samples $x \in \mathcal{M}^\ast \subset \mathbb{R}^D$ according to $x \sim {p}^\ast (x)$, where $\mathcal{M}^\ast$ is a $d$-dimensional Riemannian manifold embedded in the $D$-dimensional data space $X$, and $d \ll D$. 
We consider the two problems of estimating the density ${p}^\ast (x)$ as well as the manifold $\mathcal{M}^\ast$ given some training samples $\{x_i\} \sim {p}^\ast(x)$. 

A deep generative model represents a mapping, $g : \mathcal{Z} \rightarrow \mathcal{X}$, from a relatively low-dimensional latent space $\mathcal{Z} \subseteq \mathbb{R}^{d}$ to a high-dimensional data space $\mathcal{X} \subseteq \mathbb{R}^{D}$. 
The learned manifold $\mathcal{M}$ is a lower dimensional subset of $\mathcal{X}$, the input space of images, and is embedded in $\mathcal{Z}$ under fairly weak assumptions on the generative model itself; a generative model with a suitable capacity of representation will recover this smoothed approximation of $\mathcal{M}^\ast$.

With respect to the (constant) curvature of $\mathcal{M}$ there are three regimes of Riemannian manifolds to consider: 
Euclidean, "flat" space $\mathbb{E}^{d}$, with curvature $c = 0$;
hyperspherical, positively curved space $\mathbb{S}_{c}^{d}$, with $c > 0$;
and hyperbolic, negatively curved space $\mathbb{H}_{c}^{d}$, with $c < 0$.

By definition of Riemannian geometry, the inner-product $\langle x, x \rangle_2 = \frac{1}{c}$ for both curved regimes $\mathbb{S}^{d}$ and $\mathbb{H}^{d}$. So as $c \rightarrow 0$, both hyperspherical and hyperbolic spaces grow and become locally flatter, and $\langle x, x \rangle_2 \rightarrow \pm \infty$. This "non-convergence" property of constant curvature manifolds sends points away from the coordinate space origin in order to maintain the defined curvature of $\mathcal{M}$. 
We also observe an instability as $c \rightarrow 0$: the hyperspherical and hyperbolic geodesic distance metrics do not converge to the Euclidean distance metric.
This is an undesirable property because we \textit{a priori} must restrict the manifold curvature while learning a deep generative model.

On the other hand, stereographically projected spaces for both the hypersphere and hyperboloid manifold classes inherit the desirable properties from hyperspherical and hyperbolic spaces, while avoiding this property of sending a point to infinity when the curvature of the space has a small absolute value. This projection function
is defined as follows, for a manifold $\mathcal{M}_{k}$ of curvature $k \in \mathbb{R}$:

\vspace*{-\baselineskip}
\noindent\begin{minipage}[t]{.4\linewidth}
    {\begin{align*}
           \pi_{k} \colon \mathbb{R} \times \mathbb{R}^{n} &\to \mathcal{M}_{k}\\
           (\xi, \mathbf{x}) & \mapsto \frac{\mathbf{x}}{1 + \sqrt{|k|} \xi} 
    \end{align*}}
\end{minipage}
\begin{minipage}[t]{.6\linewidth}
    {\begin{align*}
        \pi_{k}^{-1} \colon \mathcal{M}_{k} &\to \mathbb{R} \times  \mathbb{R}^{n}\\
        \mathbf{y} & \mapsto (\frac{1}{\sqrt{|k|}} \frac{1 - k \| \mathbf{y} \|_2^2}{1 + k \| \mathbf{y} \|_2^2}, \frac{2 \mathbf{y} }{1 + k \| \mathbf{y} \|_2^2})
    \end{align*}}
\begin{equation}
\label{eq:projections}
\end{equation}
\end{minipage}
where $(\xi, x)$ is a point in the ambient space of $\mathcal{M}_{k}$, $\mathbb{R}^{n+1}$.\footnote{\citet{skopek2019mixed} similarly define such a projection function. We note our work was done concurrently, and indeed much of the findings are complimentary.}

The stereographic projections relative to the three Riemannian manifold regimes are illustrated in Fig. \ref{fig:stereo_projections}. Later we detail a novel \textit{gyroplane layer} for performing the stereographic projections in the context of a deep generative neural network.

From Eq. \ref{eq:projections} we realize several advantageous properties on these two projected spaces:

\setlength{\leftskip}{0.5cm}

\textbf{The M{\"o}bius sum} of two elements has the same structure for both projected spaces, returns an element of the same space, and only involves the Euclidean inner product (identical for every point). This has the nice consequence that the distance function in projected spaces only uses the Euclidean inner product, instead of the inner product induced by the metric tensors of the manifold, which varies on each point of the manifold. 

\textbf{The conformal projection} preserves angles, and the distance functions in the hyperbolic and hyperspherical spaces only depend on angles between vectors. This implies the hyperbolic (resp. hyperspherical) space and the Poincar{\'e} ball (resp. projected hypersphere) are isometric. 

\setlength{\leftskip}{0pt}

It is for these reasons we develop deep generative models with these two projected spaces.


If we denote $\cos_k$ the function that corresponds to $\cos$ if $k > 0$ and $\cosh$ if $k < 0$ (and similarly for $\sin_k$ and $\tan_k$), the distance function on stereographically projected spaces is:

\begin{align*}
    \Delta \mathcal{M}_{k} \left(\boldsymbol{z}_{i}, \boldsymbol{z}_{j}\right) & = \frac{1}{\sqrt{|k|}}\cos_k^{-1} \left(1+ 2 k \frac{\left\|\boldsymbol{z}_{i}-\boldsymbol{z}_{j}\right\|^{2}}{\left(1 + k \left\|\boldsymbol{z}_{i}\right\|^{2}\right)\left(1+k \left\|\boldsymbol{z}_{j}\right\|^{2}\right)}\right)
\end{align*}

whereas the gyroscopic distance function on stereographically projected spaces takes a simpler form:
\begin{equation}
    \Delta \mathcal{M}_{k} \left(\boldsymbol{z}_{i}, \boldsymbol{z}_{j}\right) = \frac{2}{\sqrt{|k|}} \tan_k^{-1} \left(\sqrt{|k|} \| - \mathbf{x} \oplus_k  \mathbf{y} \|_2\right)
\label{eq:gyro_dist}
\end{equation}
with $\oplus_k$ representing the Mobius addition.

\begin{figure}[hb]
  \centering
  {\includegraphics[width=0.9\linewidth]{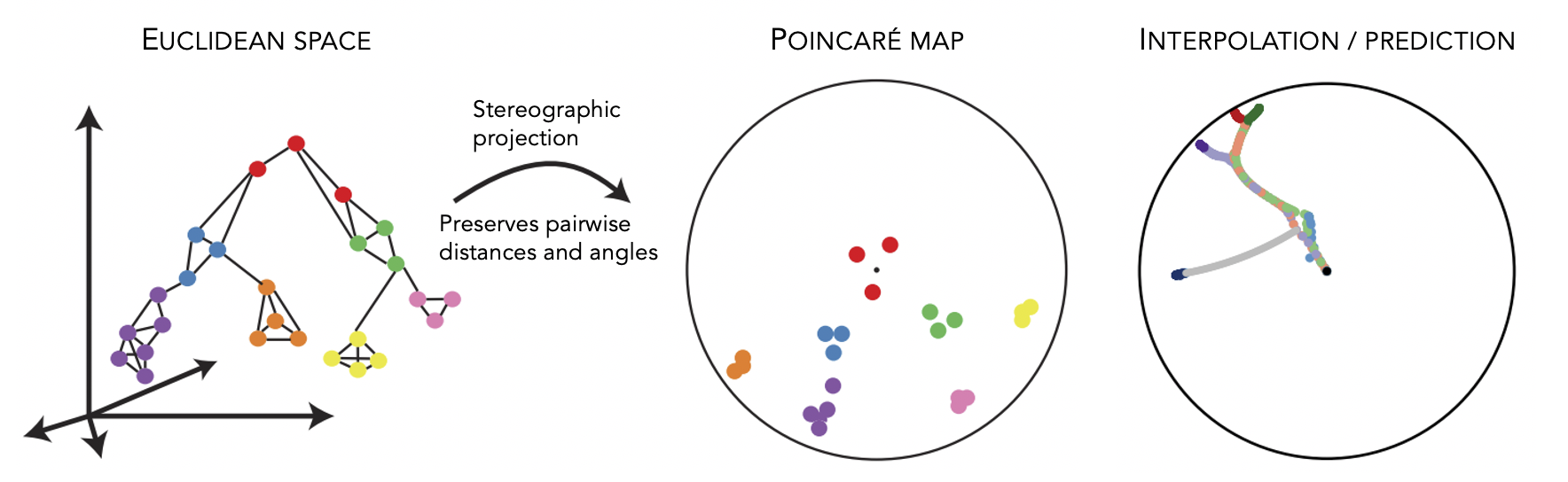}}
  \caption{
  The Poincar{\'e} ball provides meaningful geodesics for latent hierarchies, and a well-regularized space where interpolating along hyperbolic geodesics allows for reliable intermediate sampling and the prediction of unseen samples. Figure is revised from \cite{Klimovskaia2019PoincarMF}.
  }
  \label{fig:mock_poincare_map}
\end{figure}

An advantage of a smooth, regularized latent embedding space is the ability to interpolate between data points;
see Fig. \ref{fig:mock_poincare_map}. 
Interestingly, \citet{Shao2018TheRG} show straight lines in the latent space are relatively close to geodesic curves on the manifold, explaining why traversal in the latent space results in visually plausible changes to the generated data. This may work for toy datasets such as MNIST and low-quality natural images (such as CelebA faces dataset). However, in real-world images we suggest the curvilinear distances in the original data metric are not well enough preserved. Even more, we hypothesize the observations of \citet{Shao2018TheRG} will not extend beyond the standard Euclidean manifold to $c \neq 0$. 
We explore this empirically with large, complex images in histopathology datasets later.

\subsection{Manifold Learning with VAEs}

Our aim is learning manifolds for unsupervised anomaly detection. As such we focus on the Variational Auto-Encoder (VAE) \cite{Kingma2014AutoEncodingVB, Rezende2014StochasticBA} class of deep generative models. We refer the reader to our Related Work section later for treatment on comparable Generative Adversarial Networks (GANs) \cite{Goodfellow2014GenerativeAN,Radford2015UnsupervisedRL}.

The VAE is a latent variable model representing the mapping $\mathcal{Z} \rightarrow \mathcal{X}$, with an \textit{encoder} $g_{\phi}$ stochastically embedding observations in the low-dimensional latent space $\mathcal{Z}$, and a \textit{decoder} $f_{\theta}$ generating observations $\boldsymbol{x} \in \mathcal{X}$ from encodings $\boldsymbol{z} \in \mathcal{Z}$. The model uses two neural networks to respectively parameterize the likelihood $p(\cdot | f_{\theta}(\boldsymbol{z}))$ and the variational posterior $q(\cdot | g_{\phi}(\boldsymbol{x}))$.





Typically the prior distribution $p(z)$ assigned to the latent variables is a standard Gaussian. Recent work suggests this limits the capacity to learn a representative latent space (such as \cite{Mathieu2019ContinuousHR, Davidson2018HypersphericalVA}, and others discussed later in the Related Work section). We consider that the limitations of the prior are not due to a limitation in terms of capacity of representation, but more so in terms of principle. Similar to \cite{kalatzis2020variational}, we identify two major drawbacks of using the Euclidean manifold for the latent space:

\setlength{\leftskip}{0.5cm}
\paragraph{Lack of learned semantics.}
The first drawback resides in the fact that a Normal distribution or a Gaussian mixture (in a Euclidean space) can be re-parameterized in a manner that does not portray any semantic meaning for the latent data.
For instance, a mixture of Gaussians can be simply re-parameterized by a random permutation of the indices of each component in the mixture (\cite{10.5555/1162264}); while the re-parameterization is valid, the semantic meaning we associate to it is drastically different. This can rotate arbitrarily the principal components of the Euclidean latent space, and the Euclidean distance will not have a relevant meaning in terms of visual or semantic closeness in the latent space. Moreover, as has been described in \cite{arvanitidis2016locally,hauberg2018bayes}, if the decoder has a sufficient capacity of representation, it will be able to revert any re-parameterization applied in the latent space. This has the consequence that a specific value in the latent space $\mathcal{Z}$ may not be associated with a unique specific value in $\mathcal{X}$. In the context of anomaly detection, this could result in anomalous samples aligning closer to the larger groups of normal samples rather than to other anomalous samples, resulting in false negatives for entire subgroups of anomalies.

\paragraph{Irrelevant isotropic sampling.}
Secondly, \citet{Zhu_2016} suggest that human interpretable images live on a specific manifold, the "natural images manifold", noted $\mathcal{M}$ here. This manifold is a lower dimensional subset of $\mathcal{X}$, the input space of images, and is embedded in $\mathcal{Z}$ under fairly weak assumptions on the network architecture. An encoder with a suitable capacity of representation will recover a smoothed approximation of $\mathcal{M}$. This can create a latent space with a significant variable density in terms of latent samples; if our prior distribution is an isotropic Gaussian for instance, samples will be drawn in a rather isotropic manner, even though the distribution of latent samples may not have any sample in this specific area. As such, the sampling procedure in the latent space can return samples that are not relevant. Moreover, a likely consequence of this aforementioned embedding is the "manifold mismatch", or its statistical equivalent "density mismatch" \cite{Davidson2018HypersphericalVA, falorsi2018explorations}. Under the assumption of a prior distribution with an infinite support, the VAE may try to map  $\mathcal{M}$ anywhere in the space $\mathcal{Z}$, and could lead to convergence issues.

\setlength{\leftskip}{0pt}

Given the requirements of the visual anomaly detection problem, it is highly desirable to have a semantically meaningful topology which automatically embeds data according to hidden data structure, and from which we can reliably sample despite empty regions due to sparsely distributed data points. This leads us to think that an Euclidean latent space may not capture enough topological properties for visual anomaly detection.

\section{Stereographic Projections VAE}

Our aim is to construct a Poincar{\'e} ball latent space $\mathcal{Z} = \mathbb{P}_{c}^{d}$ (as shown in Fig. \ref{fig:stereo_projections}), and supporting encoder $g_{\phi}(\boldsymbol{z}|\boldsymbol{x})$ and decoder $f_{\theta}(\boldsymbol{x}|\boldsymbol{z})$ networks in order to learn a mapping from this latent space to the observation space $\mathcal{X}$.

\paragraph{Parametrising distributions on the Poincar{\'e} ball}
The choice of the probability distribution family for both the prior and the posterior (as the likelihood still lives in the Euclidean space), can be done similarly as in the Euclidean space. 
There are two distinct philosophies for adapting the Normal distribution to a Riemannian space.
The first approach is to consider the Euclidean space that is tangent at every point $z$ in the manifold, and sample from a zero-mean Euclidean Normal distribution in this tangent space.
Then, the sampled point on the tangent space is mapped to the manifold through parallel transport and the exponential map. This is known as as the "wrapping" approach. 
In the second approach, we can maximize the entropy of the distribution to derive what is known as the \textit{Riemannian Normal}.
While the latter is the only form of distribution that is proven to maximize the entropy, both distributions perform similarly in practice. Hence, we choose to use the \textit{Wrapped Normal}, as it is easier to sample from. 
We refer to both as \textit{Hyperbolic Normal} distributions with pdf $\mathcal{N}_{\mathrm{B}^{d}_{c}}\left(z | \mu, \sigma^{2}\right)$. We also define the prior on $\mathcal{Z}$ as the Hyperbolic Normal, with mean zero: $p(\boldsymbol{z}) = \mathcal{N}_{\mathrm{B}^{d}_{c}}\left(\cdot | 0, \sigma^{2}_{0}\right)$.

\paragraph{SP-VAE Architecture}
Just as in the case of a Euclidean latent space, this network is optimized by maximizing the evidence lower bound (ELBO), via an unbiased Monte Carlo (MC) estimator thanks to reparametrisable sampling schemes introduced in \cite{Mathieu2019ContinuousHR, ganea2018hyperbolic}. It was proven in \cite{Mathieu2019ContinuousHR} that the ELBO can be extended to Riemannian latent spaces by applying Jensen's inequality w.r.t. the measure on the manifold. We use $\beta$-VAE \cite{Higgins2017betaVAELB}, a variant of VAE that applies a scalar weight $\beta$ to the KL term in the objective function, as it has been shown empirically that the $\beta$-VAE improves the disentanglement of different components of the latent space when $\beta > 1$.
As we want to compare the shape of the latent manifold for visual anomaly detection in real-world applications, we chose the encoder and decoder backbones as a 4-layer convolutional network; simple enough to be able to compare all three curvature configurations, but able to learn the representation of complex images. Just as in \cite{Mathieu2019ContinuousHR,skopek2019mixed}, we use an exponential map to transform the mean of the distribution from the encoder, and then use a \textit{gyroplane layer} to go back from the Riemannian latent space to the Euclidean space, in order to take into account the shape of the manifold when applying a linear layer.

\subsection{Gyroplane Layer}

As described in \cite{ganea2018hyperbolic} and \cite{Mathieu2019ContinuousHR}, the first layer of a decoder in a VAE whose latent manifold is Euclidean and $n$-dimensional is often a linear layer. A linear layer is an affine transform, and can be written in the form
$f_{a, b} : x \mapsto \langle a, x - b \rangle$, with $x$, $a$ the orientation parameter, and $b$ the offset parameter, elements of $\mathbb{R}^n$. This expression can be rewritten as 
$f_{a, b} (x) = \sign (\langle a, x - b \rangle) \|a\| \Delta_{E} (x, H_{a, b})$, 
where $H_{a, b} = \{ x \in \mathbb{R}^n | \langle a, x - b \rangle = 0 \} = b + \{a\}^T$ is the hyperplane oriented by $a$ with offset $b$.

In the stereographically projected sphere manifold $\mathbb{D}_k^n$, the hyperplane is of the form $H_{a, p}^k = \{ z \in \mathbb{D}_k^n | \langle a, - p \oplus_k z \rangle = 0 \}$; we provide the full proof in Supplementary materials.
The distance of a point $z \in \mathbb{D}_k^n$ to $H_{a, p}^k$ takes the following form:
\begin{align}
    \Delta_k (x, H_{a, p}^k) & = \frac{1}{\sqrt{k}} \arcsin \left( \frac{2 \sqrt{k} | \langle -p \oplus_k z, a \rangle|}{(1 - k \|  -p \oplus_k z \|^2) \|a\|} \right)
\end{align}


This expression was intuitively attainable from \cite{Mathieu2019ContinuousHR,ganea2018hyperbolic}, but here we provide thorough derivation and rationale.

\section{Related Work}

\paragraph{VAE and Riemannian Manifolds} 
In Variational Auto-Encoders (VAEs) \cite{Kingma2014AutoEncodingVB, Rezende2014StochasticBA}, the prior distribution $p(z)$ assigned to the latent variables is typically a standard Gaussian. It has, unfortunately, turned out that this choice of prior is limiting the modeling capacity of VAEs and richer priors have been proposed: \citet{vampprior} propose VampPrior, a method for the latent distribution to instead be a mixture of Gaussians. \citet{vqvae} propose VQ-VAE, a way to encode more complex latent distributions with a vector quantization technique. In \cite{klushyn2019hierarchical_prior}, Klushyn et al. proposed a hierarchical prior through an alternative formulation of the objective. In \cite{bauer2018resampled}, Bauer et al. propose to refine the prior through a sampling technique.
Several notable VAEs with non-Euclidean latent spaces have been developed recently: \citet{Davidson2018HypersphericalVA} make use of hyperspherical geometry, \citet{falorsi2018explorations} endow the latent space with a SO(3) group structure, \citet{Grattarola2019AdversarialAW} introduce an adversarial auto-encoder framework with constant curvature manifold. However, in these methods the encoder and decoder are not designed to explicitly take into account the latent space geometries. Same goes for \citet{Ovinnikov2019PoincarWA}, who proposed to use a Poincar{\'e} ball latent space, but were not able to derive a closed-form solution of the ELBO’s entropy term. \citet{Mathieu2019ContinuousHR} propose the Poincar{\'e} VAE, closely aligned with our work. We extend it mainly to consider practical properties of the manifold geometries towards real applications, arriving at the stereographic projection mechanisms. The method most related to the current paper is mixed-curvature VAE from \citet{skopek2019mixed}. They similarly define a projection across hyperboloid and hypersphere spaces for use in VAEs. Our work was done concurrently, and much of the findings are complimentary.

\paragraph{Visual Anomaly Detection}
Anomaly detection is a deep field with many application areas in machine learning. We focus on the image domain, referring the reader to \citet{Chandola2009AnomalyDA, Pimentel2014ARO} and references therein for full surveys of the field. 
A promising area in visual anomaly detection is reconstruction-based methods, with recent works that train deep autoencoders to detect anomalies based on reconstruction error \cite{Zhou2017AnomalyDW, Zhai2016DeepSE, Zong2018DeepAG}. For example, \citet{Zhai2016DeepSE} use a structured energy based deep neural network to model the training samples, and \citet{Zong2018DeepAG} proposed to jointly model the encoded features and the reconstruction error in a deep autoencoder. Although the reconstruction-based methods have shown promising results, their performances are ultimately restricted by the under-designed representation of the latent space.
While we focus on images, there exist methods for videos such as applying PCA with optical flow methods \cite{Kim2009ObserveLI} and RNNs for next-frame predictions \cite{Luo2017ARO}.
\citet{Schlegl2017UnsupervisedAD} applied Generative Adverserial Networks (GANs) to the task of VAD. Their AnoGAN was succeeded my the more efficient EGBAD that uses a BiGAN approach \cite{Zenati2018EfficientGA}. In \cite{Akay2018GANomalySA},the combination of a GAN and autoencoder was introduced.
For more on GANs in anomaly detection please refer to \cite{Mattia2019ASO}.
We compare against GANs in the Experiments section, and show superior results with our VAE method. Even more, VAEs are a preferable class of deep generative models because they provide a natural probabilistic formulation, readily work with various priors, and are easier to train.

\section{Experiments}


\subsection{Visual Anomaly Detection Problem Setup}

In this paper we consider two related but distinct problems of unsupervised anomaly detection in images: scoring and localization. 
Let $\mathcal{X}$ be the space of all images in our domain of interest, and let $X \subseteq \mathcal{X}$ be the set of images defined as normal. 
We investigate two different metrics: the reconstruction error probability, which can be used for both tasks, as well as the ELBO derivative with respect to the input. For scoring, we use the average value ($\mu_{rec}$) and standard deviation ($\sigma_{rec}$) of the reconstruction error on all pixels on the test set, and take a threshold at $\mu_{rec} + 1.5\sigma_{rec}$. The producing a mask from this result gives us the anomaly localization.



We evaluated our approach on several benchmark datasets for visual anomaly detection. Importantly, we focus on those with real-world images.
Prior works limit evaluations to MNIST and Omniglot datasets, which are not representative of natural images. 

\subsection{Crack Segmentation \& PCB Defects Benchmarks}
\label{sec:experiments}

We experiment on two benchmark anomaly detection datasets of real-world images. The Crack Segmentation dataset contains images of cracked surfaces (brick walls, concrete roads, lumpy surfaces, etc.), concatenating images from several datasets: Crack 500 \cite{zhang2016road}, CrackTree200 \cite{zou2012cracktree} and AELLT \cite{Amhaz2016AutomaticCD}, and others.\footnote{Crack Segmentation dataset is available at \url{github.com/khanhha/crack_segmentation}.}
We also experiment with the PCB Dataset \cite{huang2019pcb} for defect detection in precision manufacturing. The dataset contains 3597 training images, 1161 validation images, and 1148 testing images, at various resolutions. The dataset is made from defect-free images, and defects are added in images with annotations, including positions of the six most common types of PCB defects (open, short, mousebite, spur, pin hole, and spurious copper). 


\begin{figure}[!h]
  \centering
  {\includegraphics[width=0.65\linewidth]{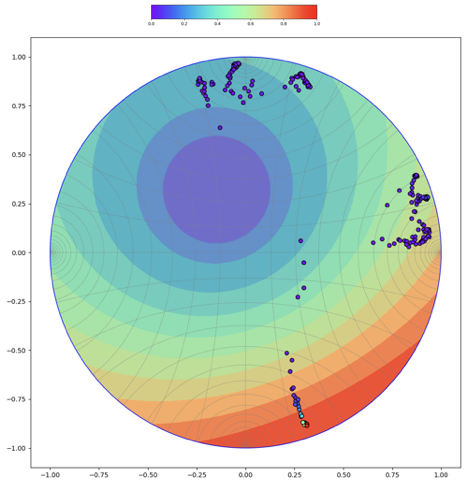}}
  \caption{2-D Poincar{\'e} Ball Embeddings for the PCB dataset, with SVDD scores level lines. Purple points are normal instances, others are anomalous instances.}
  
  \label{fig:anomaly_ball}
\end{figure}

In constructing the PCB dataset, PCB images were divided in non-overlapping $128 \times 128$ patches. Patches that contain anomalous pixels where stored in the anomalous set of patches and the remaining patches in the "normal" set $\mathcal{D}_{normal}$. Due to the creation process for this dataset, the "anomaly" set has a larger amount of elements than the "normal" set. In order to fit usual real industrial inspection data distribution, a random subset of the anomalous samples was selected, with $n_{anomaly} = 0.1 \vert \mathcal{D}_{normal} \vert$.

In order to obtain the Poincar{\'e} embeddings for the PCB dataset, we adapted unsupervised anomaly detection method Deep SVDD \cite{Ruff2018DeepOC} for the Poincar{\'e} ball latent space. The Auto-Encoder (AE) used a ResNet-18 (\cite{resnet}) backbone for the encoder, followed by a $\log_0^k$-map and two hyperbolic linear layers to obtain feature vectors in the Poincar{\'e} ball manifold; the latent space is 2-dimensional. The decoder was composed of an $\exp_0^k$-operator followed by a deconvolutional ResNet-18 as the backbone. The AE was pretrained for $350$ epochs, with the Riemannian Adam optimizer from the Geoopt library (\cite{geoopt2020kochurov}), and a learning rate of $10^{-4}$ in the first $250$ epochs and $10^{-5}$ for the remaining epochs. 

The second step of the Deep SVDD method starts with center-initialization. In the Euclidean case, the initialization is accomplished by averaging all the features vectors from the training set as output by the trained encoder. In the hyperbolic case, with feature vectors on the Poincar{\'e} ball, we computed the gyrobarycenter instead of the Euclidean average. Then, the encoder was fine-tuned with the following loss: 

\begin{equation}
    \mathcal{L} = \sum_{i=1}^n \Delta_k (x, c)^2 + \lambda \sum_{k=1}^W \|w_k\|^2
\end{equation}

with $c$, the initialized center, $\{w_k\}_{k \in \{0, \dots, W\}}$ the weights of the encoder, and $\lambda = 5\times 10^{-7}$ the $L^2$ regularization parameter. 
The encoder was then trained with the SVDD objective, formulated for the Poincar{\'e} ball manifold, also with the Riemannian Adam optimizer and the optimization parameters from the original paper for 150 epochs.

Anomaly scores were computed as: $d_k (\phi (x), c)$, with $x$ an input sample, $c$ the center as defined in our previous gryrobarycenter calculation, and $\phi$ the mapping learning by the encoder.  The radius of the hyperbolic "sphere" was selected similarly to the Deep SVDD paper, as the $90\%$-ile of the the computed scores on the testing set.  All samples whose score exceeded this radius were classified as anomalies, as shown in Fig. \ref{fig:anomaly_ball}.
We applied the hyperbolic UMAP algorithm (with $n_{neighbors}=50$ and $dist_{min} = 0.001$) to produce easily interpretable figures for the Poincar{\'e} embeddings, with level lines of the anomaly score function inside the ball.

Below are results for both visual anomaly detection tasks, on these two datasets:

\begin{table}[h!]
\small
\captionsetup{font=small}
    \centering
    \begin{tabular}{|c|c|c|c|c|}
    \hline
        Datasets            & Precision & Recall & F1 & IoU \\ \hline
         Crack segmentation &  0.4206& 1.0 &  0.5921&   0.5470\\ \hline
         PCB                &  0.4514 & 0.9228  & 0.6063  & 0.2942 \\ \hline
    \end{tabular}
    \caption{Euclidean, dimension 6}
\end{table}
\vspace*{-\baselineskip}
\begin{table}[h!]
\small
\captionsetup{font=small}
    \centering
    \begin{tabular}{|c|c|c|c|c|}
    \hline
        Datasets                    & Precision & Recall & F1 & IoU \\ \hline
         Crack segmentation &  0.4205 & 1.0 & 0.4205  & 0.5083 \\ \hline
         PCB                         &  0.4462 &  0.9520 &  0.6076 &  0.2911\\ \hline
    
    \end{tabular}
    \caption{Projected Sphere, dimension 6}
\end{table}
\vspace*{-\baselineskip}
\begin{table}[h!]
\small
\captionsetup{font=small}
\centering
    \centering
    \begin{tabular}{|c|c|c|c|c|}
    \hline
    Datasets                    & Precision & Recall & F1       & IoU \\ \hline
        Crack segmentation      &  0.42055  & 0.9994 & 0.59199  & 0.5087   \\ \hline
         PCB                    & 0.4264    & 0.9530   & 0.5801    &  0.2950\\ \hline
     \end{tabular}
    \caption{Poincar{\'e} Ball, dimension 6}
\end{table}
\vspace*{-\baselineskip}

Overall, all three manifolds perform similarly across datasets. On images with very orthogonal features, the Euclidean manifold seems to perform better, both on the scoring and segmentation tasks. For crack images, which contain very non-linear cracks, the Poincar{\'e} ball performs best for the localization tasks.
\subsection{Application in Histopathology}


\begin{figure}[!h]
  \centering
  {\includegraphics[width=0.7\linewidth]{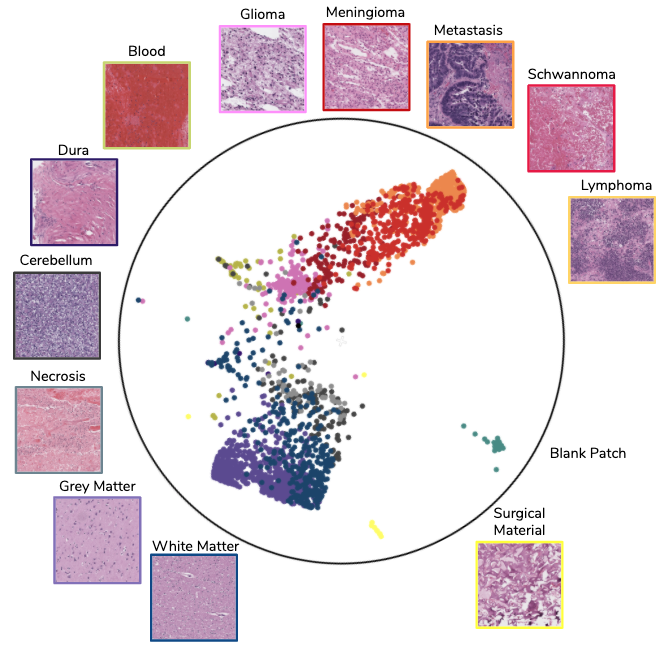}}
  \caption{Poincar{\'e} ball of the learned latent embedding, showing a structure that separates cancerous tissue (top) from normal tissue (bottom) and non-tissue (e.g. surgical material). We also see some semantically meaningful hierarchy develop; the manifold center splits normal and ab-normal tissues, and progressing down the branch of cancerous tissue we see patterns such as cohesive lesions (meningioma and metastasis) being arranged close together. Importantly this organization is learned unsupervised. 
  Figure is best viewed in color.}
\label{fig:latent_brains}
\end{figure}

We investigated the applicability of our approach to the challenging task of diagnostic neuropathology, the branch of pathology focused on the microscopic examination of neurosurgical specimens. 
We experimented with a dataset of H\&E-stained whole-slide images (WSI) of a glioblastoma containing a heterogeneous mixture of tumor, necrosis, brain tissue, blood and surgical material. See the Supplement for dataset details.

Manual inspection of WSI to sufficiently search for metastatic cells amongst normal cells is infeasible; a single WSI may contain millions of lymphocytes, nucleotides, and other cells for inspection. 
Automated and unsupervised computer vision methods could prove invaluable. 
Even more, the task of diagnosis can be error-prone and subjective. 
For one, overlapping patterns of the most common brain tumor types -- gliomas, meningiomas, schwannomas, metastases, and lymphomas -- present a challenge. 
Even more, although these five tumor types represent the majority of cases encountered in clinical practice ($\sim$75–80\%), there are over 100 different brains tumor subtypes to be considered, many of which are exceedingly rare \cite{cnsReport2014}. Similarly, new diseases (e.g. Zika encephalitis) continually arise.
For these reasons, we hypothesize the latent hierarchical representation learned by our Stereographic Projection VAE can delineate these complex subtypes while providing a continuous trajectory relating any two points.

The experiment setup was as follows: 
We trained unsupervised on a dataset of 1024 x 1024 pixel images tiled from WSIs, representing eight non-lesional categories (hemorrhage, surgical material, dura, necrosis, blank slide space, and normal cortical gray, white, and cerebellar brain tissue), and the aforementioned five common lesional subtypes (gliomas, meningiomas, schwannomas, metastases, and lymphomas). 
Fig. \ref{fig:latent_brains} shows example patches from each of these categories, displayed across the learned hyperbolic manifold.
The Supplement additionally contains WSI examples and data details.

The model used here is the same as in the experiments described earlier, but for a slight change in the $\beta$ parameter: We used a simulated annealing approach to progressively update $\beta$ as a function of the reconstruction loss
during training. 
Some methods simply use a linear increase schedule for $\beta$, but such predefined schedules may be suboptimal.
The method is not unlike that of \citet{Klushyn2019LearningHP}, 
for which we provide details in the Supplementary materials.


We find that our model learns a latent hierarchical embedding that organizes the distributions of normal tissue, non-tissue materials, and cancerous tissue, shown in Fig. \ref{fig:latent_brains}. 
Not to mention the manifold embedding reveals interpretable semantics of known and potentially unknown tissue relationships.
In the context of unsupervised visual anomaly detection, we learn this latent embedding without labels, and identify cancerous tissues as sparse anomalies that are distributed on Poincar{\'e} ball regions opposite normal tissue.
If we then train a classifier on the anomalous samples that are discretized by the unsupervised embedding, we achieve a classification performance of > 0.97 across the five disease subtypes, as assessed by the areas under the multi-class receiver operator curve (AUC, mROC); this is consistent with the classification scheme and state-of-the-art results in the fully supervised approach of \cite{Faust2018VisualizingHD}).

\begin{figure}[!tbp]
  \centering
  {\includegraphics[width=0.95\linewidth]{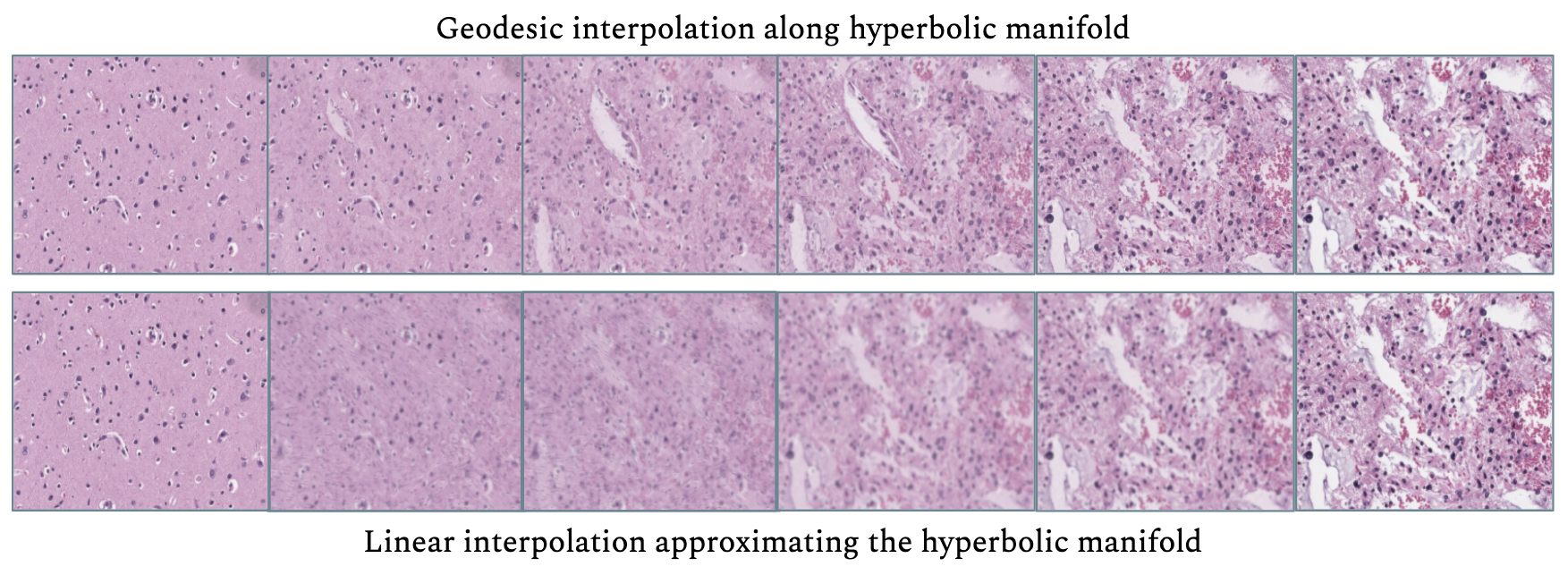}}
  \caption{
  Interpolation along the learned manifold from normal grey matter brain tissue (left) to cancerous glioma tissue (right). The top row represents samples along the geodesic path, while the bottom represents samples from the linear approximation.
  We find that interpolating along the curved manifold yields intermediary samples that are reasonable tissue images, perhaps representing true intermediate cellular states. Linear interpolation comparatively yields blurred intermediary structures.
  For example, notice the center-top white gap structure that grows organically as the geodesic samples progress from normal to cancerous. Comparatively, the linear samples on the bottom row show unnatural blending.
  Note both rows are sub-sampled from the latent space learned by our Stereographic Projection VAE, but along different interpolation paths. The Euclidean latent space of a $\beta$-VAE yielded blurry samples without cellular structures.
  }
  \label{fig:tissue_interpolation}
\end{figure}

As mentioned earlier, interpolation in the latent space of a Euclidean VAE is possible because, for simple data regimes, the linear interpolation metric closely approximates the true geodesic curves. We suggest this approximation does not necessarily hold when using a non-Euclidean latent space, particularly when the deep generative model is learning in a complex image space where curvature plays a more prominent role.
We investigate this by carrying out geodesic interpolations and comparing these with the corresponding linear counterparts in $\mathcal{Z}$ space. We use Eqn. \ref{eq:gyro_dist} to estimate the geodesic curve connecting a given pair of images on the generated manifold, discretizing the curve at 10 points. To get an image on the generated manifold, we pick a real image $x$ from the dataset and use $g(h(x))$ to get the corresponding point on the generated manifold. Fig. \ref{fig:tissue_interpolation} shows example linear and geodesic interpolations between the same endpoints on $\mathcal{Z}$. 
We find the linear approximations to be unreliable, counter to prior findings that focused on Euclidean manifolds \cite{Shao2018TheRG} and relatively simple images.

Even more, we suggest the linear approximation can yield calibration errors when probing for specific points along an interpolation "arc", particularly with long-range interpolations between sparse data points that arise in anomaly detection settings. We elucidate this in Fig. \ref{fig:tissue_interpolation}.
In scientific endeavors such as interpolating between known and rare brain tumor representations, we need points that reliably lie on the manifold surface such that generated images represent plausible samples.

\section{Conclusion}

In this paper we explored Riemannian manifolds in deep generative neural networks towards unsupervised visual anomaly detection. Key insights were derived from investigations into specific properties of Riemannian curvatures that best enable natural and efficient embedding of both normal data and sparse anomalous data.
To work with such manifolds in the context of Variational Auto-Encoders (VAEs), we derived a gyroplane layer that enables stereographic projections between hyperspherical and hyperbolic latent spaces: a \textit{Stereographic Projection VAE}.
Empirically we found our hypotheses to be valid, and matched state-of-the-art results on real world benchmarks.
We also made valuable observations regarding manifold interpolations and sampling, finding linear approximations of geodesic curves to be unreliable.
In the challenging domain of neuropathology, our model learns a latent hierarchical organization of brain cancer subtypes and other tissues, despite not using labels. 
Using \textit{Poincar{\'e} mapping} we effectively interpolate across the manifold, yielding reliable intermediate samples from the manifold.
Without this capability, a deep generative model does not necessarily satisfy the \textit{manifold hypothesis for natural images}.
Future work would be to continue development of our approach in histopathology, 
as this can be a valuable decision support tool for pathologists, and can theoretically be applied to diverse tissue and disease classes.
We would also like to continue working with Poincar{\'e} mapping and other methods that can derive insights directly from the rich latent space.

\bibliographystyle{plainnat}
\bibliography{vad_neurips2020}

\begin{thebibliography}{53}
\providecommand{\natexlab}[1]{#1}
\providecommand{\url}[1]{\texttt{#1}}
\expandafter\ifx\csname urlstyle\endcsname\relax
  \providecommand{\doi}[1]{doi: #1}\else
  \providecommand{\doi}{doi: \begingroup \urlstyle{rm}\Url}\fi

\bibitem[Akçay et~al.(2018)Akçay, Abarghouei, and
  Breckon]{Akay2018GANomalySA}
Samet Akçay, Amir~Atapour Abarghouei, and Toby~P. Breckon.
\newblock Ganomaly: Semi-supervised anomaly detection via adversarial training.
\newblock \emph{ArXiv}, abs/1805.06725, 2018.

\bibitem[Alemi et~al.(2018)Alemi, Poole, Fischer, Dillon, Saurous, and
  Murphy]{Alemi2018FixingAB}
Alexander~A. Alemi, Ben Poole, Ian~S. Fischer, Joshua~V. Dillon, Rif~A.
  Saurous, and Kevin Murphy.
\newblock Fixing a broken elbo.
\newblock In \emph{ICML}, 2018.

\bibitem[Amhaz et~al.(2016)Amhaz, Chambon, Idier, and
  Baltazart]{Amhaz2016AutomaticCD}
Rabih Amhaz, Sylvie Chambon, J{\'e}r{\^o}me Idier, and Vincent Baltazart.
\newblock Automatic crack detection on two-dimensional pavement images: An
  algorithm based on minimal path selection.
\newblock \emph{IEEE Transactions on Intelligent Transportation Systems},
  17:\penalty0 2718--2729, 2016.

\bibitem[An and Cho(2015)]{An2015VariationalAB}
Jinwon An and Sungzoon Cho.
\newblock Variational autoencoder based anomaly detection using reconstruction
  probability.
\newblock 2015.

\bibitem[Arvanitidis et~al.(2016)Arvanitidis, Hansen, and
  Hauberg]{arvanitidis2016locally}
Georgios Arvanitidis, Lars~Kai Hansen, and Søren Hauberg.
\newblock A locally adaptive normal distribution, 2016.

\bibitem[Bauer and Mnih(2018)]{bauer2018resampled}
Matthias Bauer and Andriy Mnih.
\newblock Resampled priors for variational autoencoders, 2018.

\bibitem[Bishop(2006)]{10.5555/1162264}
Christopher~M. Bishop.
\newblock \emph{Pattern Recognition and Machine Learning (Information Science
  and Statistics)}.
\newblock Springer-Verlag, Berlin, Heidelberg, 2006.
\newblock ISBN 0387310738.

\bibitem[Chandola et~al.(2009)Chandola, Banerjee, and
  Kumar]{Chandola2009AnomalyDA}
Varun Chandola, Arindam Banerjee, and Vipin Kumar.
\newblock Anomaly detection: A survey.
\newblock \emph{ACM Comput. Surv.}, 41:\penalty0 15:1--15:58, 2009.

\bibitem[Davidson et~al.(2018)Davidson, Falorsi, Cao, Kipf, and
  Tomczak]{Davidson2018HypersphericalVA}
Tim~R. Davidson, Luca Falorsi, Nicola~De Cao, Thomas Kipf, and Jakub~M.
  Tomczak.
\newblock Hyperspherical variational auto-encoders.
\newblock In \emph{UAI}, 2018.

\bibitem[Deecke et~al.(2018)Deecke, Vandermeulen, Ruff, Mandt, and
  Kloft]{Deecke2018ImageAD}
Lucas Deecke, Robert~A. Vandermeulen, Lukas Ruff, Stephan Mandt, and Marius
  Kloft.
\newblock Image anomaly detection with generative adversarial networks.
\newblock In \emph{ECML/PKDD}, 2018.

\bibitem[et~al.(2014)]{cnsReport2014}
Quinn~Ostrom et~al.
\newblock Cbtrus statistical report: primary brain and central nervous system
  tumors diagnosed in the united states in 2007-2011.
\newblock \emph{Neuro Oncol}, 16 Suppl 4:\penalty0 iv1--63, 2014.

\bibitem[Falorsi et~al.(2018)Falorsi, de~Haan, Davidson, Cao, Weiler, Forré,
  and Cohen]{falorsi2018explorations}
Luca Falorsi, Pim de~Haan, Tim~R. Davidson, Nicola~De Cao, Maurice Weiler,
  Patrick Forré, and Taco~S. Cohen.
\newblock Explorations in homeomorphic variational auto-encoding, 2018.

\bibitem[Faust et~al.(2018)Faust, Xie, Han, Goyle, Volynskaya, Djuric, and
  Diamandis]{Faust2018VisualizingHD}
Kevin Faust, Quin Xie, Dominick Han, Kartikay Goyle, Zoya~I. Volynskaya,
  Ugljesa Djuric, and Phedias Diamandis.
\newblock Visualizing histopathologic deep learning classification and anomaly
  detection using nonlinear feature space dimensionality reduction.
\newblock \emph{BMC Bioinformatics}, 19, 2018.

\bibitem[Ganea et~al.(2018)Ganea, Bécigneul, and Hofmann]{ganea2018hyperbolic}
Octavian-Eugen Ganea, Gary Bécigneul, and Thomas Hofmann.
\newblock Hyperbolic neural networks, 2018.

\bibitem[Goodfellow et~al.(2014)Goodfellow, Pouget-Abadie, Mirza, Xu,
  Warde-Farley, Ozair, Courville, and Bengio]{Goodfellow2014GenerativeAN}
Ian~J. Goodfellow, Jean Pouget-Abadie, Mehdi Mirza, Bing Xu, David
  Warde-Farley, Sherjil Ozair, Aaron~C. Courville, and Yoshua Bengio.
\newblock Generative adversarial nets.
\newblock In \emph{NIPS}, 2014.

\bibitem[Grattarola et~al.(2019)Grattarola, Livi, and
  Alippi]{Grattarola2019AdversarialAW}
Daniele Grattarola, Lorenzo Livi, and Cesare Alippi.
\newblock Adversarial autoencoders with constant-curvature latent manifolds.
\newblock \emph{Appl. Soft Comput.}, 81, 2019.

\bibitem[Haney and Lavin(2020)]{Haney2020HypersphereVision}
Bijan Haney and Alexander Lavin.
\newblock Fine-grain few-shot vision via domain knowledge as hyperspherical
  priors.
\newblock In \emph{CVPR Workshop on Fine-grained Categorization}, 2020.

\bibitem[Hauberg(2018)]{hauberg2018bayes}
Søren Hauberg.
\newblock Only bayes should learn a manifold (on the estimation of differential
  geometric structure from data), 2018.

\bibitem[He et~al.(2015)He, Zhang, Ren, and Sun]{resnet}
Kaiming He, Xiangyu Zhang, Shaoqing Ren, and Jian Sun.
\newblock Deep residual learning for image recognition.
\newblock \emph{CoRR}, abs/1512.03385, 2015.
\newblock URL \url{http://arxiv.org/abs/1512.03385}.

\bibitem[Higgins et~al.(2017)Higgins, Matthey, Pal, Burgess, Glorot, Botvinick,
  Mohamed, and Lerchner]{Higgins2017betaVAELB}
Irina Higgins, Lo{\"i}c Matthey, Arka Pal, Christopher Burgess, Xavier Glorot,
  Matthew~M Botvinick, Shakir Mohamed, and Alexander Lerchner.
\newblock beta-vae: Learning basic visual concepts with a constrained
  variational framework.
\newblock In \emph{ICLR}, 2017.

\bibitem[Huang and Wei(2019)]{huang2019pcb}
Weibo Huang and Peng Wei.
\newblock A pcb dataset for defects detection and classification, 2019.

\bibitem[Kalatzis et~al.(2020)Kalatzis, Eklund, Arvanitidis, and
  Hauberg]{kalatzis2020variational}
Dimitris Kalatzis, David Eklund, Georgios Arvanitidis, and Søren Hauberg.
\newblock Variational autoencoders with riemannian brownian motion priors,
  2020.

\bibitem[Kim and Grauman(2009)]{Kim2009ObserveLI}
Jaechul Kim and Kristen Grauman.
\newblock Observe locally, infer globally: A space-time mrf for detecting
  abnormal activities with incremental updates.
\newblock In \emph{CVPR}, 2009.

\bibitem[Kingma and Welling(2014)]{Kingma2014AutoEncodingVB}
Diederik~P. Kingma and Max Welling.
\newblock Auto-encoding variational bayes.
\newblock \emph{CoRR}, abs/1312.6114, 2014.

\bibitem[Klimovskaia et~al.(2019)Klimovskaia, Lopez-Paz, Bottou, and
  Nickel]{Klimovskaia2019PoincarMF}
Anna Klimovskaia, David Lopez-Paz, L{\'e}on Bottou, and Maximilian Nickel.
\newblock Poincar{\'e} maps for analyzing complex hierarchies in single-cell
  data.
\newblock \emph{bioRxiv}, 2019.

\bibitem[Klushyn et~al.(2019{\natexlab{a}})Klushyn, Chen, Kurle, Cseke, and
  van~der Smagt]{Klushyn2019LearningHP}
Alexej Klushyn, Nutan Chen, Richard Kurle, Botond Cseke, and Patrick van~der
  Smagt.
\newblock Learning hierarchical priors in vaes.
\newblock \emph{ArXiv}, abs/1905.04982, 2019{\natexlab{a}}.

\bibitem[Klushyn et~al.(2019{\natexlab{b}})Klushyn, Chen, Kurle, Cseke, and
  van~der Smagt]{klushyn2019hierarchical_prior}
Alexej Klushyn, Nutan Chen, Richard Kurle, Botond Cseke, and Patrick van~der
  Smagt.
\newblock Learning hierarchical priors in vaes, 2019{\natexlab{b}}.

\bibitem[Kochurov et~al.(2020)Kochurov, Karimov, and
  Kozlukov]{geoopt2020kochurov}
Max Kochurov, Rasul Karimov, and Serge Kozlukov.
\newblock Geoopt: Riemannian optimization in pytorch, 2020.

\bibitem[Luo et~al.(2017)Luo, Liu, and Gao]{Luo2017ARO}
Weixin Luo, Wen Liu, and Shenghua Gao.
\newblock A revisit of sparse coding based anomaly detection in stacked rnn
  framework.
\newblock \emph{2017 IEEE International Conference on Computer Vision (ICCV)},
  pages 341--349, 2017.

\bibitem[Mathieu et~al.(2019)Mathieu, Lan, Maddison, Tomioka, and
  Teh]{Mathieu2019ContinuousHR}
Emile Mathieu, Charline~Le Lan, Chris~J. Maddison, Ryota Tomioka, and Yee~Whye
  Teh.
\newblock Continuous hierarchical representations with poincar{\'e} variational
  auto-encoders.
\newblock In \emph{NeurIPS}, 2019.

\bibitem[Mattia et~al.(2019)Mattia, Galeone, Simoni, and Ghelfi]{Mattia2019ASO}
Federico~Di Mattia, Paolo Galeone, Michele~D. Simoni, and Emanuele Ghelfi.
\newblock A survey on gans for anomaly detection.
\newblock \emph{ArXiv}, abs/1906.11632, 2019.

\bibitem[Mettes et~al.(2019)Mettes, van~der Pol, and
  Snoek]{Mettes2019HypersphericalPN}
Pascal Mettes, Elise van~der Pol, and Cees G.~M. Snoek.
\newblock Hyperspherical prototype networks.
\newblock In \emph{NeurIPS}, 2019.

\bibitem[Nickel and Kiela(2017)]{Nickel2017PoincarEF}
Maximilian Nickel and Douwe Kiela.
\newblock Poincar{\'e} embeddings for learning hierarchical representations.
\newblock \emph{ArXiv}, abs/1705.08039, 2017.

\bibitem[Ovinnikov(2019)]{Ovinnikov2019PoincarWA}
Ivan Ovinnikov.
\newblock Poincar{\'e} wasserstein autoencoder.
\newblock \emph{ArXiv}, abs/1901.01427, 2019.

\bibitem[Pidhorskyi et~al.(2018)Pidhorskyi, Almohsen, Adjeroh, and
  Doretto]{Pidhorskyi2018GenerativePN}
Stanislav Pidhorskyi, Ranya Almohsen, Donald~A. Adjeroh, and Gianfranco
  Doretto.
\newblock Generative probabilistic novelty detection with adversarial
  autoencoders.
\newblock In \emph{NeurIPS}, 2018.

\bibitem[Pimentel et~al.(2014)Pimentel, Clifton, Clifton, and
  Tarassenko]{Pimentel2014ARO}
Marco A.~F. Pimentel, David~A. Clifton, Lei~A. Clifton, and Lionel Tarassenko.
\newblock A review of novelty detection.
\newblock \emph{Signal Process.}, 99:\penalty0 215--249, 2014.

\bibitem[Radford et~al.(2015)Radford, Metz, and
  Chintala]{Radford2015UnsupervisedRL}
Alec Radford, Luke Metz, and Soumith Chintala.
\newblock Unsupervised representation learning with deep convolutional
  generative adversarial networks.
\newblock \emph{CoRR}, abs/1511.06434, 2015.

\bibitem[Rezende et~al.(2014)Rezende, Mohamed, and
  Wierstra]{Rezende2014StochasticBA}
Danilo~Jimenez Rezende, Shakir Mohamed, and Daan Wierstra.
\newblock Stochastic backpropagation and approximate inference in deep
  generative models.
\newblock In \emph{ICML}, 2014.

\bibitem[Ruff et~al.(2018)Ruff, G{\"o}rnitz, Deecke, Siddiqui, Vandermeulen,
  Binder, M{\"u}ller, and Kloft]{Ruff2018DeepOC}
Lukas Ruff, Nico G{\"o}rnitz, Lucas Deecke, Shoaib~Ahmed Siddiqui, Robert~A.
  Vandermeulen, Alexander Binder, Emmanuel M{\"u}ller, and Marius Kloft.
\newblock Deep one-class classification.
\newblock In \emph{ICML}, 2018.

\bibitem[Schlegl et~al.(2017)Schlegl, Seeb{\"o}ck, Waldstein, Schmidt-Erfurth,
  and Langs]{Schlegl2017UnsupervisedAD}
Thomas Schlegl, Philipp Seeb{\"o}ck, Sebastian~M. Waldstein, Ursula
  Schmidt-Erfurth, and Georg Langs.
\newblock Unsupervised anomaly detection with generative adversarial networks
  to guide marker discovery.
\newblock In \emph{IPMI}, 2017.

\bibitem[Shao et~al.(2018)Shao, Kumar, and Fletcher]{Shao2018TheRG}
Hang Shao, Abhishek Kumar, and P.~Thomas Fletcher.
\newblock The riemannian geometry of deep generative models.
\newblock \emph{2018 IEEE/CVF Conference on Computer Vision and Pattern
  Recognition Workshops (CVPRW)}, pages 428--4288, 2018.

\bibitem[Skopek et~al.(2019)Skopek, Ganea, and B{\'e}cigneul]{skopek2019mixed}
Ondrej Skopek, Octavian-Eugen Ganea, and Gary B{\'e}cigneul.
\newblock Mixed-curvature variational autoencoders.
\newblock \emph{arXiv preprint arXiv:1911.08411}, 2019.

\bibitem[Tifrea et~al.(2019)Tifrea, Becigneul, and Ganea]{tifrea2018poincare}
Alexandru Tifrea, Gary Becigneul, and Octavian-Eugen Ganea.
\newblock Poincare glove: Hyperbolic word embeddings.
\newblock In \emph{International Conference on Learning Representations}, 2019.

\bibitem[Tomczak and Welling(2017)]{vampprior}
Jakub~M. Tomczak and Max Welling.
\newblock {VAE} with a vampprior.
\newblock \emph{CoRR}, abs/1705.07120, 2017.
\newblock URL \url{http://arxiv.org/abs/1705.07120}.

\bibitem[Ungar(2009)]{gyro_book}
Abraham Ungar.
\newblock \emph{A Gyrovector Space Approach to Hyperbolic Geometry}, volume~1.
\newblock 01 2009.
\newblock \doi{10.2200/S00175ED1V01Y200901MAS004}.

\bibitem[van~den Oord et~al.(2017)van~den Oord, Vinyals, and
  Kavukcuoglu]{vqvae}
Aaron van~den Oord, Oriol Vinyals, and Koray Kavukcuoglu.
\newblock Neural discrete representation learning, 2017.

\bibitem[Zenati et~al.(2018)Zenati, Foo, Lecouat, Manek, and
  Chandrasekhar]{Zenati2018EfficientGA}
Houssam Zenati, Chuan~Sheng Foo, Bruno Lecouat, Gaurav Manek, and
  Vijay~Ramaseshan Chandrasekhar.
\newblock Efficient gan-based anomaly detection.
\newblock \emph{ArXiv}, abs/1802.06222, 2018.

\bibitem[Zhai et~al.(2016)Zhai, Cheng, Lu, and Zhang]{Zhai2016DeepSE}
Shuangfei Zhai, Yu~Cheng, Weining Lu, and Zhongfei Zhang.
\newblock Deep structured energy based models for anomaly detection.
\newblock \emph{ArXiv}, abs/1605.07717, 2016.

\bibitem[Zhang et~al.(2016)Zhang, Yang, Zhang, and Zhu]{zhang2016road}
Lei Zhang, Fan Yang, Yimin~Daniel Zhang, and Ying~Julie Zhu.
\newblock Road crack detection using deep convolutional neural network.
\newblock In \emph{Image Processing (ICIP), 2016 IEEE International Conference
  on}, pages 3708--3712. IEEE, 2016.

\bibitem[Zhou and Paffenroth(2017)]{Zhou2017AnomalyDW}
Chong Zhou and Randy~C. Paffenroth.
\newblock Anomaly detection with robust deep autoencoders.
\newblock \emph{Proceedings of the 23rd ACM SIGKDD International Conference on
  Knowledge Discovery and Data Mining}, 2017.

\bibitem[Zhu et~al.(2016)Zhu, Krähenbühl, Shechtman, and Efros]{Zhu_2016}
Jun-Yan Zhu, Philipp Krähenbühl, Eli Shechtman, and Alexei~A. Efros.
\newblock Generative visual manipulation on the natural image manifold.
\newblock \emph{Lecture Notes in Computer Science}, page 597–613, 2016.

\bibitem[Zong et~al.(2018)Zong, Song, Min, Cheng, Lumezanu, ki~Cho, and
  Chen]{Zong2018DeepAG}
Bo~Zong, Qi~Song, Martin~Renqiang Min, Wei Cheng, Cristian Lumezanu, Dae
  ki~Cho, and Haifeng Chen.
\newblock Deep autoencoding gaussian mixture model for unsupervised anomaly
  detection.
\newblock In \emph{ICLR}, 2018.

\bibitem[Zou et~al.(2012)Zou, Cao, Li, Mao, and Wang]{zou2012cracktree}
Qin Zou, Yu~Cao, Qingquan Li, Qingzhou Mao, and Song Wang.
\newblock Cracktree: Automatic crack detection from pavement images.
\newblock \emph{Pattern Recognition Letters}, 33\penalty0 (3):\penalty0
  227--238, 2012.

\end{thebibliography}

\appendix
\section*{Supplementary Materials
}
\renewcommand{\thesubsection}{\Alph{subsection}}

\subsection{Notations}

\begin{itemize}
    \item $(H, W) \in \mathbb{N}^2$: height and width of input images
    \item $\mathcal{X}$: input image space, included in $\mathbb{R}^{H\times W}$, considered Euclidean
    \item $Z \in \mathbb{N}$: latent dimension
    \item $\mathcal{Z}$: latent space, included in $\mathbb{R}^{Z}$, with $Z \ll  H\times W$
    \item $p_{data} : \mathcal{X} \rightarrow [0, 1]$: probability distribution of the data
    \item $x \sim p_{data}$: random variable draw in input space
    \item $p_z :\mathcal{Z} \rightarrow [0, 1] $ prior probability
    \item $p_{\phi}(z | x)$: posterior probability distribution, learned by the encoder
    \item $p_{\theta}(x | z)$: likelihood distribution, learned by the decoder
\end{itemize}









\subsection{Gyroplane layer derivation}
\label{supp:gyro}

Here we provide the proof of the gyroplane layer; it is similar to the one in \cite{ganea2018hyperbolic}, with the following expression:

The distance of a point $z \in \mathbb{D}_k^n$ to $H_{a, p}^k$ takes the form:
\begin{align}
    \Delta_k (x, H_{a, p}^k) & = \frac{1}{\sqrt{k}} \arcsin \left( \frac{2 \sqrt{k} | \langle -p \oplus_k z, a \rangle|}{(1 - k \|  -p \oplus_k z \|^2) \|a\|} \right)
    \label{eq:to-prove}
\end{align}

\subsubsection{A few definitions}
Here are some definitions of notions we are going to use in the proof. They come from Riemannian geometry and \cite{gyro_book}.

$\mathbb{D}_k^n$ is a $n$-dimensional manifold. The tangent space at a point $z \in \mathbb{D}_k^n$ is noted $\mathcal{T}_z \mathbb{D}_k^n$. The Riemannian metric $g = (g_z)_{z \in \mathbb{D}_k^n}$ is a set of inner products $g_z : \mathcal{T}_z \mathbb{D}_k^n \times \mathcal{T}_z \mathbb{D}_k^n \rightarrow \mathbb{R}$, varying in a smooth manner with $z$. 

\begin{definition}[Mobius addition]
The \textbf{Mobius addition} is defined as follows, for two points $x, y \in \mathbb{D}_k^n$:
\begin{align}
    x \oplus_{k} y = \frac{(1 - 2k \langle x, y \rangle -k \|y\|^2) x + (1 + k \|x\|^2)y}{1 - 2k \langle x, y \rangle + k^2 \|x\|^2 \|y\|^2}
\end{align}
\end{definition}

It is worth noting that this addition is neither commutative nor associative, but it does have the following properties, $\forall x \in \mathbb{D}_k^n$:
\begin{itemize}
    \item $x \oplus_{k} 0_{\mathbb{D}_k} = 0_{\mathbb{D}_k} \oplus_{k} x = x$
    \item $(-x) \oplus_{k} x = x \oplus_{k} (-x) = 0_{\mathbb{D}_k}$
    \item $\forall y \in \mathbb{D}_k$, $(-x) \oplus_{k} (x \oplus_{k} y) = y$ (left cancellation law)
\end{itemize}
The \textbf{Mobius subtraction} is simply defined as $x \ominus_k y = x \oplus_{k} (-y)$.

\begin{definition}[Gyroangle]
For $x, y, z \in \mathbb{D}_k^n$, we will denote by $\hat{yxz}$ the angle between the two geodesics starting from $x$ and ending at $y$ and $z$ respectively. This angle, named the \textit{gyroangle} can be defined either by the angle between the two initial velocities of each geodesic, $u$ and $v$:
\begin{align}
    \cos{(\widehat{(u,v)})} & = \frac{\langle u, v \rangle}{\|u\| \|v\|}
\end{align}
Or as:
\begin{align}
    \cos{(\widehat{yxz})} & = \langle \frac{-x \oplus_k y}{\|-x \oplus_k y\|}, \frac{-x \oplus_k z}{\|-x \oplus_k z\|} \rangle
\end{align}
\end{definition}

\begin{definition}
The \textbf{Gyrodistance} between two points $x, y \in \mathbb{D}_k^n $ is defined as: $d_{\oplus_k}(x, y) = \|y \ominus_k x\|$.
\end{definition}

\begin{definition}
In \cite{gyro_book}, Ungar defined a \textbf{Gyroline} as: $\forall x, y \in \mathbb{D}_k^n $, $\forall t \in [0, 1]$, $\gamma (t) = x \oplus_k (\ominus_k x \oplus_k y) \otimes_k t$, where $\otimes_k$ is the \textbf{Mobius Scalar mutliplication} in the Gyrogroup $(\mathbb{D}_k^n, \oplus_k)$, defined as:

\begin{align}
    t \otimes_k v &= \frac{((1 + \sqrt{k}\|v\|)^t - (1 -  \sqrt{k}\|v\|)^t)}{ ((1+ \sqrt{k} \|v\|)^t +(1 - \sqrt{k} \|v\|)^t )} \times \frac{v}{\|v\|} \\
                  &= \frac{1}{\sqrt{k}} \tan (t \arctan (\sqrt{k} \|v\|)) \frac{v}{\|v\|}
\end{align}
\end{definition}

So the geodesic $\gamma_{x \rightarrow y} (t) = x \oplus_k (\ominus_k x \oplus_k y) \otimes_k t$, with $\gamma_{x \rightarrow y} : \mathbb{R} \rightarrow \mathbb{D}_k^n$, that satisfies the following constraints: $\gamma_{x \rightarrow y} (0) = x$ and $\gamma_{x \rightarrow y} (1) = y$. If we use this definition, and do a reparametrization using the gyrodistance definition to make this geodesic of constant speed, we obtain that the unit speed geodesic starting at $x \in \mathbb{D}_k^n $ with direction $v \in \mathcal{T}_x \mathbb{D}_k^n$ is:

\begin{align}
    \gamma_{x, v}(t) = x \oplus_k \left(\tan{\sqrt{k} \frac{t}{2}} \frac{v}{\sqrt{k} \|v\|}\right)
\end{align}

with $\gamma_{x, v} : \mathbb{R} \rightarrow \mathbb{D}_k^n$, that satisfies the following constraints: $\gamma_{x, v} (0) = x$ and $\dot{\gamma}_{x, v} (0) = v$.


\begin{definition}[Log Map]
The Log Map of the stereographically projected sphere is defined, for $x$ and $y$ in $\mathbb{D}_k^n$:
\begin{align}
    \log_x^k (y) &= \frac{2}{\sqrt{k} \lambda_x^k} \arctan (\sqrt{k} \|-x \oplus_k y\|) \frac{-x \oplus_k y}{\|-x \oplus_k y\|}
\end{align}
\end{definition}

\begin{definition}[Exp Map]
The Log Map of the stereographically projected sphere is defined, for $x \in \mathbb{D}_k^n$ and $v \in \mathcal{T}_x\mathbb{D}_k^n$:
\begin{align}
    \exp_x^k (v) &= x \oplus_k \tan (\sqrt{k} \frac{\lambda_x^k \|v\|}{2} \frac{v}{\sqrt{k} \|v\|})
\end{align}
\end{definition}

\subsubsection{Stereographically projected sphere hyperplane}
We are defining what an hyperplane in the stereographically projected sphere; the final expression needs justification, so the proof follows the definition.
\begin{definition}[Stereographically projected sphere hyperplane]

For a point $p \in \mathbb{D}_k^n$, a point $a \in \mathcal{T}_{p} \mathbb{D}_k^n \setminus \{0\}$, let $\{a\}^{\perp} = \{ z \in \mathcal{T}_{p} \mathbb{D}_k^n | g_k^p (z, a) = 0\}$. Since $g_k^p (z, a) = {\lambda_x^{k}}^2 \langle z, a \rangle$ and $\lambda_p^{k} = \frac{2}{1 + k \| p \|^2} > 0 , \  \forall p \in \mathbb{D}_k^n$, we have that: $\{a\}^{\perp} = \{ z \in \mathcal{T}_{p} \mathbb{D}_k^n | \langle z, a\rangle = 0\}$. The hyperplane in the stereographically projected sphere is defined as:

\begin{align}
    H_{a,p} &= \{z \in  \mathcal{T}_{p} \mathbb{D}_k^n | \langle z, a\rangle = 0 \} \nonumber \\
            &= \{x \in \mathbb{D}_k^n \setminus {p}| \langle \log_p^k (x), a\rangle_p = 0 \} \bigcup \{p\} \label{eq1} \\
            &= \{x \in \mathbb{D}_k^n | \langle -p \oplus_k x, a \rangle = 0\} \label{eq3}
\end{align}
\end{definition}
\begin{proof}
If $z = p$, $\langle z, a \rangle = 0$ by definition of the tangent space.

If $z \neq p$, since $\arctan$ is a strictly increasing function on $ ]-\pi / 2, \pi / 2[  \rightarrow \mathbb{R}$, it is a bijection from $\mathbb{R}$ to $[-\pi /2 , \pi /2]$. For a fixed $p \in \mathbb{D}_k^n$, $f \colon x \mapsto -p \oplus_k x$ is also a bijection so $\forall z \in \mathcal{T}_{p} \mathbb{D}_k^n , \exists ! x \in \mathbb{D}_k^n \ s.t. \ z = \log_p^k (x)$, which proves equality \ref{eq1}. 

Still in the case of $z \neq p$:
\begin{align}
    \langle \log_p^k (x), a\rangle_p = 0 & \Leftrightarrow \frac{2}{\sqrt{k} \lambda_p^k} \arctan{\sqrt{k} \|-p \oplus_k x \|} \langle \frac{-p \oplus_k x }{\|-p \oplus_k x \|}, a \rangle = 0 \label{eq:log_map} \\
                                         & \Leftrightarrow \arctan{(\sqrt{k} \|-p \oplus_k x \|)} = 0 \ \text{or} \  \langle \frac{-p \oplus_k x }{\|-p \oplus_k x \|}, a \rangle = 0 \label{eq:atan}\\
                                         & \Leftrightarrow \langle \frac{-p \oplus_k x }{\|-p \oplus_k x \|}, a \rangle = 0 \label{eq:inner_prod}
\end{align}
\ref{eq:log_map} is obtained by definition of the logarithm map of $\mathbb{D}_k^n$. Since $\frac{2}{\sqrt{k} \lambda_p^k} > 0$ , we obtain \ref{eq:atan}. Finally, since $\sqrt{k}\|-p \oplus_k x \| > 0 $ ($x \neq p$), then $\arctan{(\sqrt{k}\|-p \oplus_k x \|)} > 0$, so we obtain \ref{eq:inner_prod}.
Lastly, if $x = p$, \ref{eq3} is still true, which achieves the proof. 
\end{proof}

\subsubsection{Distance to hyperplane}

We are going to proceed in four steps: firstly, we need to prove the existence and unicity of the orthogonal projection of a point in the manifold on a geodesic that does not go through this point. Then, we will prove that this projection in fact minimizes the distance between the point and the geodesic. Then, we will prove that geodesics that pass through 2 points of an hyperplane belong entirely to that hyperplane, and finally, we will find the explicit expression of this distance.

\paragraph{Existence and Unicity of an orthogonal projection on a geodesic}
\label{sec:ex-uni}

Thanks to the preliminary theorem, we know that the orthogonal projection of a point $x \in \mathbb{D}_k^n$ on a given geodesic $\gamma$ of $ \mathbb{D}_k^n$ that does not include $x$ exists and is unique. 
\paragraph{Minimizing distance between a point and a geodesic}
This projection minimizes the distance between the point $x$ and the geodesic $\gamma$, since the hypotenuse in a spherical right triangle is strictly longer than the two other sides of the rectangle (constant curvature space sine law).   

\paragraph{Geodesics in $H_{a, p}^k$}

Let $H_{a, p}^k$ be an hyperplane in the stereographically projected sphere. 
Let $w$ be a point in $H_{a, p}^k$, such that $w \neq p$. Let us consider the geodesic $\gamma_{p \rightarrow w}$ (it exists since $w \neq p$). As we have previously seen in \ref{}, this geodesic is of the form, $\forall t \in \mathbb{R}$:
\begin{align}
\gamma_{p \rightarrow w} (t) = p \oplus_k (-p \oplus_k w) \otimes_k t 
\end{align}
 We want to see if, $\forall t \in \mathbb{R}$, $\gamma_{p \rightarrow w} (t) $ belongs to $H_{a, p}^k$, that is to say: $\langle -p \oplus_k \gamma_{p \rightarrow w} (t), a \rangle = 0$ :
 
 \begin{align}
 	\langle -p \oplus_k \gamma_{p \rightarrow w} (t), a \rangle &= \langle (-p \oplus_k w) \otimes_k t, a \rangle \label{eq:left-cancellation} \\ 
	                                                                                            &= \frac{1}{\sqrt{k}} \tan (t \arctan (\sqrt{k} \|-p \oplus_k w \|)) \langle -p \oplus_k w, a \rangle \label{eq:def-scalar-prod} \\
	                                                                                            &= 0 \label{eq:eq3-hp}
 \end{align}
 We obtain \ref{eq:left-cancellation} by the use of the left cancellation law; then, by definition of the M\"obius scalar product, we obtain \ref{eq:def-scalar-prod}. And we finally obtain \ref{eq:eq3-hp} since $w \in H_{a, p}^k$. 
 
 \paragraph{Distance to hyperplane expression}
 
 Let $x$ be a point in $\mathbb{D}_k^n$, and $H_{a, p}^k$ be an hyperplane in $\mathbb{D}_k^n$. Let us denote $w^*$ the point that minimizes $d_k (x, w)$ (it exists since $d_k$ is continuous in both variables and bounded below). For a $w \in H_{a, p}^k$,  then $x$, $w$ and $p$ form a gyrotriangle in $\mathbb{D}_k^n$, noted $\Delta xwp$.  Let us suppose now that $\widehat{xwp} \neq = \pi / 2$, then $w \neq w^*$ (by \ref{sec:ex-uni}). 
 
 From now on, in order to find $w^*$, we are hence going to consider points $w$ such that $\widehat{xwp} = \pi / 2$ (we know $w^*$ in the set $\{w \in H_{a, p}^k \vert  \widehat{xwp} \neq \pi / 2 \} $). By applying the constant curvature sine law in the right triangle $\Delta xwp$, we obtain:

 \begin{equation}
d_k (x, w) = \frac{1}{\sqrt{k}} \arcsin (\sin (\sqrt{k}d_k (x, p)) \sin ( \widehat{xwp}))
\end{equation}
But we have that:
 \begin{equation}
\sin (\sqrt{k}d_k (x, p)) = \sin (2 \arctan (\sqrt{k} \|-p \oplus_k x\|))
\end{equation}
 
 By using the trigonometric property : $\sin (2x) = \frac{2 \tan (x)}{ 1 + \tan^2 (x)}$, we obtain that:
 
   \begin{equation}
\sin (\sqrt{k}d_k (x, p)) = \frac{2 \sqrt{k} \|-p \oplus_k x \|}{1 + k \|-p \oplus_k x\|^2}
\end{equation}
and does not depend on $w$.

The term inside the $\arcsin$ that influences the minimization of $d_k (x, w)$ is $\sin( \widehat{xwp})$. Since $\sin (x) = \sqrt{1 - \cos^2 (x)}$, minimizing $\sin( \widehat{xwp})$ is equivalent to maximizing $\cos( \widehat{xwp})$, to which we know the expression: 

\begin{align}
\cos (\widehat{xwp}) &= \frac{\langle -p \oplus_k x,  -p \oplus_k w\rangle}{\| -p \oplus_k x\| \|-p \oplus_k w\|}
\end{align}

It is quite straightforward to prove that $\{ \log_p^k (w) \vert w \in H_{a, p}^k\} = \{ a\}^{\perp}$. Since $\log_p^k (w) = \frac{2}{\sqrt{k} \lambda_p^k} \arctan (\sqrt{k} \|-p \oplus_k w\|) \frac{-p \oplus_k w}{\|-p \oplus_k w\|}$, we have that: $\|\log_p^k (w) \|= \| \frac{2}{\sqrt{k} \lambda_p^k} \arctan (\sqrt{k} \|-p \oplus_k w\|) \|$. Since $\sqrt{k} \|-p \oplus_k w\| > 0$, then $ \arctan (\sqrt{k} \|-p \oplus_k w\|) > 0$. We know that $\frac{2}{\sqrt{k} \lambda_p^k} > 0$, so the real number $ \| \frac{2}{\sqrt{k} \lambda_p^k} \arctan (\sqrt{k} \|-p \oplus_k w\|) \| =  \frac{2}{\sqrt{k} \lambda_p^k} \arctan (\sqrt{k} \|-p \oplus_k w\|)$. We can deduce that:

\begin{align}
\frac{\log_p^k (w)}{\|\log_p^k (w)\|} &= \frac{-p \oplus_k w}{\|-p \oplus_k w\|}
\end{align}

And our optimization problem becomes:

\begin{align}
\max_{z \in \{a\}^{\perp}} & \frac{\langle -p \oplus_k x, z \rangle}{\|-p \oplus_k x \| \|z\|}
\end{align}

Which is a Euclidean optimization problem, whose solution is well known; we obtain:

\begin{align}
\sin (\widehat{xpw^*}) &= \frac{\vert \langle -p \oplus_k x, a \rangle \vert }{\|-p \oplus_k x\| \|a\|}
\end{align}
So, $w^*$ exists by construction, and by re-injecting the expression of $\sin (\widehat{xpw^*})$ in $d_k (x, w^*)$, we obtain the expression in \ref{eq:to-prove}.



\subsection{Datasets details}
\label{supp:data}

\paragraph{Neuropathology} We us the digitized brain tissue dataset as described in \cite{Faust2018VisualizingHD}: Brain tissue slides were digitized into whole-slide images (WSI), as shown in Fig. \ref{fig:tissue_interpolation}. Each WSI is tiled into an image patch of 1024 x 1024 pixels (0.504 $\mu m$ per pixel, 516 $\mu m^2$) to carry out training and inference, a tile size over 10 times larger than most other approaches. This larger size was chosen because it contains multiple levels of morphologic detail (single cell-level and overall tumor structure) without significantly affecting computation times. The size of WSI varies, most with length and width of approximately 50,000 pixels. In the construction of the dataset by \citet{Faust2018VisualizingHD}, all samples were anonymized and annotations carried out by board-certified pathologists. Slides were annotated with eight non-lesional categories (hemorrhage, surgical material, dura, necrosis, blank slide space, and normal cortical gray, white, and cerebellar brain tissue), and five common lesional subtypes (gliomas, meningiomas, schwannomas, metastases, and lymphomas). The training image dataset and WSI testing cases are available for download in the Zenodo repository at \url{https://doi.org/10.5281/zenodo.1237976} and \url{https://doi.org/10.5281/zenodo.1238084}, respectively. 
Below in Table S1 we breakdown the tissue types in the dataset.

\begin{figure}[!h]
  \centering
  {\includegraphics[width=0.95\linewidth]{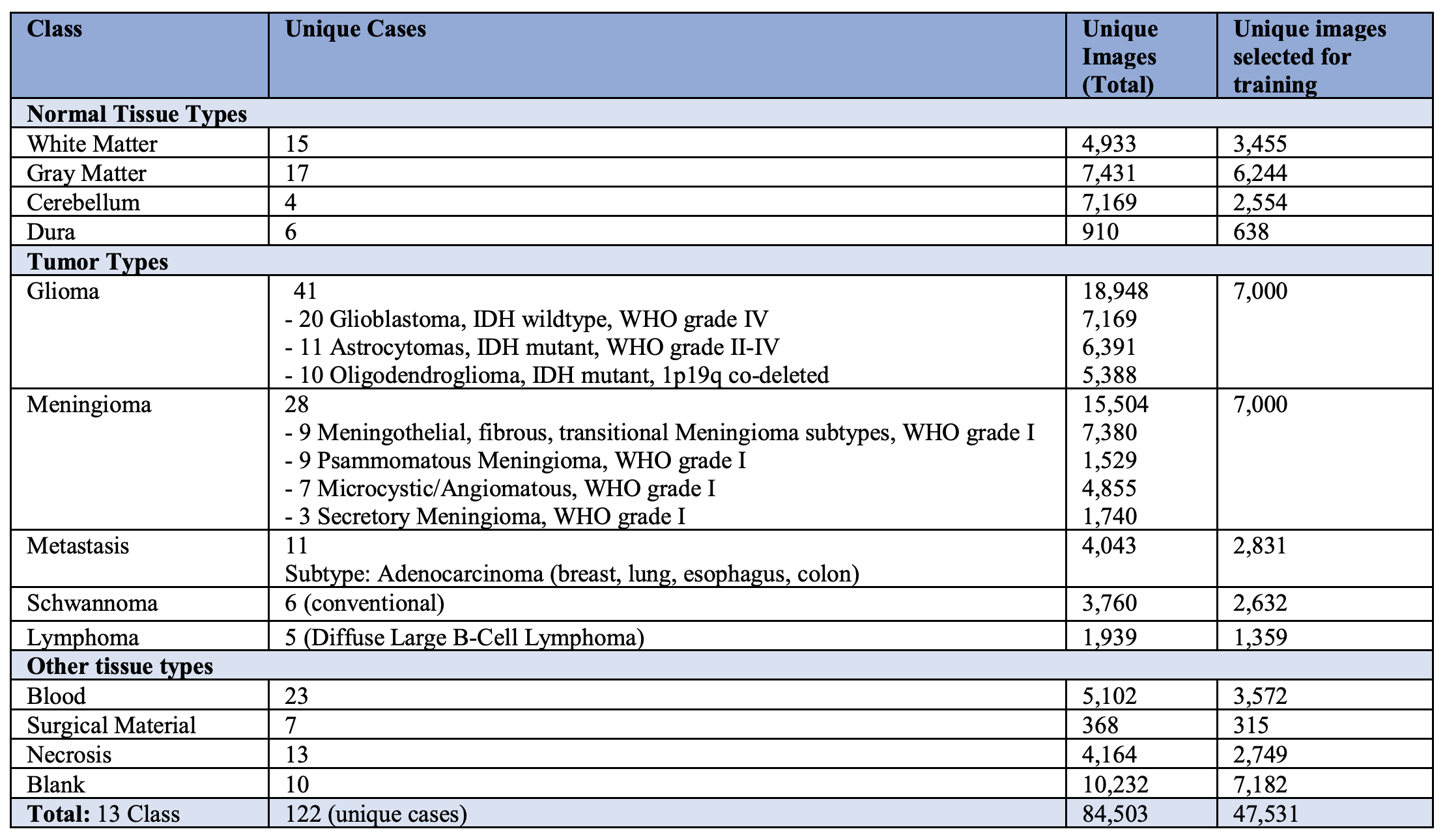}}
  \\ \small Table S1. Distribution of tissue types and images used for training. These numbers represent the aforementioned 1024 x 1024 pixel images. Reproduced from \cite{Faust2018VisualizingHD}.
  \label{fig:histopath_training_dataset}
\end{figure}

\begin{figure}[!t]
  \centering
  {\includegraphics[width=0.7\linewidth]{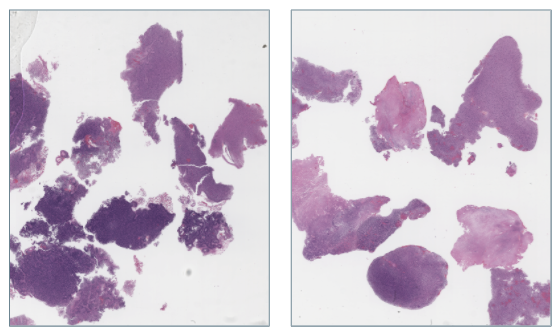}}
  \caption{Two example H\&E-stained whole-slide images (WSI) of glioblastoma from the neuropathology test set, each containing a heterogeneous mixture of tumor, necrosis, brain tissue, blood, and surgical material. Each WSI contains several thousand 1024 x 1024 tiles, as shown in Fig. \ref{fig:latent_brains}.}
  \label{fig:wsi_examples}
\end{figure}

\subsection{Model hyperparameters and setup}
\label{supp:models}

\textbf{TODO (Louise)}: params for the models used in experiments

For all experiments we use $\beta$-VAE \cite{Higgins2017betaVAELB}, a variant of VAE that applies a scalar weight $\beta$ to the KL term in the objective function. In the histopathology experiments we found stronger reconstruction results with a weighting schedule applied to the KL term of the ELBO. This is because a different ratio targets different regions in the rate-distortion plane, either favouring better compression or reconstruction \cite{Alemi2018FixingAB}.

We start with $\beta \ll 1$ to enforce a reconstruction optimization. When the average reconstruction error $C_{\theta}(\boldsymbol{x},\boldsymbol{z})$ hits a predefined parameter $\kappa^2$ we initiate the following update scheme:

\begin{align}
    \beta_t = \beta_{t-1} \cdot exp(\nu \cdot (\hat{C}_{t} - \nu^2))
\end{align}

where $\nu$ is the update's learning rate.

For the experiments in section \ref{sec:experiments}, the backbone encoder consisted of 4-convolutional layers (2D Convolution + Batch Normalization + Leaky ReLU), with a hidden Euclidean dimension of $400$. The optimization was done with the Adam optimizer, with a constant learning rate of $10^-4$, and a batch size of $128$. The maximum number of epochs was set to $250$, with an early stopping mechanism, with $150$ warm-up epochs and $80$ epochs for the lookahead. 

\end{document}